# Robust UAV Position and Attitude Estimation using Multiple GNSS Receivers for Laser-based 3D Mapping

Taro Suzuki[1], Daichi Inoue[2], Yoshiharu Amano[2]

*Abstract*— Small-sized unmanned aerial vehicles (UAVs) have been widely investigated for use in a variety of applications such as remote sensing and aerial surveying. Direct three-dimensional (3D) mapping using a small-sized UAV equipped with a laser scanner is required for numerous remote sensing applications. In direct 3D mapping, the precise information about the position and attitude of the UAV is necessary for constructing 3D maps. In this study, we propose a novel and robust technique for estimating the position and attitude of small-sized UAVs by employing multiple low-cost and light-weight global navigation satellite system (GNSS) antennas/receivers. Using the "redundancy" of multiple GNSS receivers, we enhance the performance of real-time kinematic (RTK)-GNSS by employing single-frequency GNSS receivers. This method consists of two approaches: hybrid GNSS fix solutions and consistency examination of the GNSS signal strength. The fix rate of RTK-GNSS using single-frequency GNSS receivers can be highly enhanced to combine multiple RTK-GNSS to fix solutions in the multiple antennas. In addition, positioning accuracy and fix rate can be further enhanced to detect multipath signals by using multiple GNSS antennas. In this study, we developed a prototype UAV that is equipped with six GNSS antennas /receivers. From the static test results, we conclude that the proposed technique can enhance the accuracy of the position and attitude estimation in multipath environments. From the flight test, the proposed system could generate a 3D map with an accuracy of 5 cm.

## I. Introduction

Small-sized unmanned aerial vehicles (UAVs) are less expensive and more convenient to operate than manned aircrafts, and they can also fly at low altitudes to acquire high-resolution data. However, performing 3D mapping using small-sized UAVs poses several challenges. The strict weight limit of small-sized UAVs imposes restrictions on the size and weight of on-board equipment such as sensors. This restriction in turn limits the on-board installation of instruments such as laser scanners or light detection and ranging (LiDAR) used in aerial surveys as well as precise position and attitude estimation systems that are based on a dual-frequency global navigation satellite system (GNSS) and a high-grade inertial measurement unit (IMU). Owing to the above-mentioned constraints, which prevent the use of high-precision UAV position and attitude sensors, it is challenging to estimate the accurate position and attitude of small-sized UAVs, which is essential for 3D mapping.

Two approaches are available for constructing 3D maps: indirect 3D mapping, and direct 3D mapping [1, 2]. Figure 1 illustrates the dissimilarity between these two approaches. In both the approaches, 3D-map generation requires accurate position and attitude data of the UAV to transform the coordinates of the relative sensor data obtained from the UAV. However, in indirect 3D mapping, a method of simultaneously estimating both the position and attitude data of a UAV, and a 3D map is used in a framework called the simultaneous localization and mapping (SLAM) or the structure-from-motion technique [3-7]. A camera is typically used for constructing the 3D map, and numerous optimization-based techniques that use bundle-adjustment technique [3] have been proposed. In this case, the position and attitude of a UAV are estimated from sequential images or video recordings obtained using a camera installed on the UAV; thus, the precise position and attitude data of the UAV are not required for 3D mapping. Several software tools developed for construction of 3D UAV maps from aerial images [6] have already been used in industry and science applications [7-8]. However, the disadvantage of indirect 3D mapping is that it requires field surveys to include geographical information (such as latitude, longitude, and altitude) in the generated 3D point clouds. To analyze the generated 3D maps, a high-quality georeferencing performance is required. Moreover, observing ground control points (GCPs) or natural landmarks (LMs) from images is essential for georeferencing in indirect 3D mapping. In addition, accurate GCP or LM locations must be predetermined by the prior field surveys. Thus, the indirect 3D mapping approach is challenging to use in disaster-struck environments that are inaccessible to humans. LiDAR-based odometry and the SLAM techniques have been developed and used in ground mobile robots. However, in the case of the UAV laser survey, typically, UAVs fly at 60--80 m altitude to construct 3D terrain map; thus, continuous features cannot be obtained from the data obtained using LiDAR-based odometry and the SLAM technique. Furthermore, LiDAR-based odometry and the SLAM technique are difficult to apply in estimating UAV position and attitude.

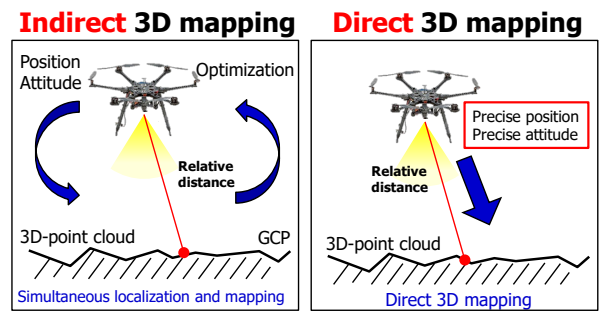

Figure 1. Dissimilarities between indirect and direct 3D mapping approaches using UAVs. In the case of direct 3D mapping with a laser scanner, the precise position and attitude data are required.

[1]Taro Suzuki is with the Future Robotics Technology Center, Chiba Institute of Technology, Japan (e-mail: taro@furo.org)

[2]Daichi Inoue and Yoshiharu Amano are with Waseda University, Japan.

In contrast, in the case of direct 3D mapping, a UAV's accurate position and attitude data must be directly measured using on-board sensors installed on the UAV. In this case, a laser scanner rather than a camera is required for generating the 3D map because a monocular camera cannot directly measure the relative distance of an object with respect to the UAV and can only measure the direction of the object. If a flying UAV's precise position and attitude data can be measured using on-board sensors, then 3D point clouds are automatically generated from the laser-scanned data on the basis of the position and attitude data of the UAV. The main advantage of using direct 3D mapping during flight is that no field surveys are required to determine GCPs for performing georeferencing. Practically, a direct 3D mapping system can be used in unknown environments such as disaster-struck environments and environments inaccessible to humans.

The real-time kinematic (RTK)-GNSS positioning technology can be used to estimate the UAV's position with centimeter-level accuracy. In general, a dual-frequency GNSS receiver is used in the RTK-GNSS technology. A few studies have employed a dual-frequency GNSS receiver to estimate UAV position, and a high-grade gyroscope (e.g., fiber optic gyroscope and ring laser gyroscope) to estimate UAV's attitude [9-10]. However, these receivers are not practical when considering their cost, size, and weight, which, in most of the cases, surpasses the payload limit of small-sized UAVs.

On the other hand, RTK-GNSSs that use low-cost and light-weight single-frequency GNSS receivers and antennas are attracting significant attention in numerous applications. In recent years, many GNSSs have been launched in several countries, and the number of positioning satellites has also increased rapidly. This increase in the number of positioning satellites has enabled the implementation of single-frequency RTK-GNSSs. However, using low-cost single-frequency GNSS receivers in the single-frequency RTK-GNSS technique results in higher degradation of the carrier-phase ambiguity resolution as compared to that obtained using dual-frequency GNSS receivers. In particular, the performance of RTK-GNSSs has considerably degraded in urban environments that contain numerous buildings, because of the reflection and diffraction of signals (also known as multipath signals) caused due to the buildings [11]. In UAV applications regarding the inspection of structures or use in urban and mountainous areas, it is challenging to estimate the UAV's position with high precision by using a low-cost single-frequency GNSS receiver.

This study aims to establish a method for accurately estimating the attitude and position of small-sized UAVs, by using only low-cost GNSS receivers. The key concept behind the proposed method is the utilization of multiple low-cost single-frequency GNSS antennas/receivers to accurately estimate the attitude and position of a UAV. Thus, the performance of the RTK-GNSS is enhanced by utilizing the "redundancy" of multiple GNSS receivers.

*A Contribution*

We developed a novel method and a system to robustly estimate the precise position and attitude of a UAV by using multiple GNSS antennas/receivers. We developed the method to enhance the fix ratio of single-frequency RTK-GNSS to combine multiple GNSS-based solutions. In addition, we proposed a new technique to estimate the GNSS multipath signal on the basis of the observation of multiple GNSS receivers to enhance the RTK-GNSS's performance. Furthermore, we achieved the 3D-mapping accuracy of 5 cm using low-cost multiple GNSS receivers and a laser scanner, without using IMUs.

II. PROPOSED SYSTEM

*A. Unmanned Aerial Vehicle*

We developed a distinctive multicopter equipped with six single-frequency GNSS antennas/receivers to estimate the precise position and attitude of a UAV. Figure 2 illustrates the developed multicopter. The GNSS antennas were installed on the exterior of the UAV's propellers by extending the arm of each propeller. The maximum distance between the antennas is approximately 1.8 m. We used low-cost u-blox M8T GNSS receivers, which can output GNSS carrier-phase measurements for RTK-GNSS and Harxon HX-CH3602 GNSS antennas. We used Velodyne VLP-16 LiDAR for performing 3D mapping. The LiDAR can synchronize the receivers' data with GPS-supplied time pulses. The six GNSS receivers were synchronized with their GPS timestamps. The problem regarding sensor time synchronization does not occur in the proposed system. The total weight of all the installed equipment was 1813 g, and the multicopter system can be used in numerous small-sized UAV systems.

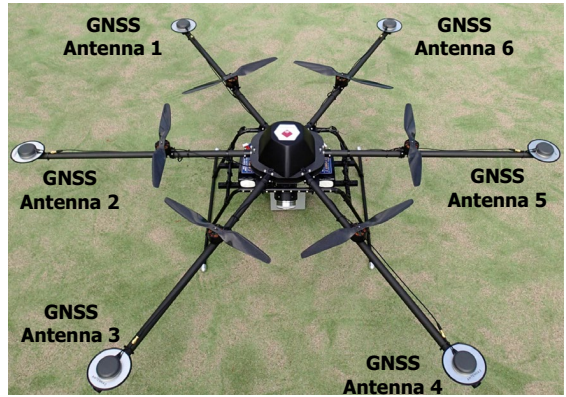

Figure 2. Multicopter using six GNSS antennas/receivers.

*B. Overview of the Proposed Method*

The key concept behind this study is the use of RTK-GNSS with multiple low-cost and single-frequency GNSS receivers/antennas to enhance the accuracy of a UAV's position and attitude estimation. In general, only one GNSS receiver is sufficient for estimating the UAV's position, and if we use three GNSS receivers, two baseline vectors can be estimated between GNSS antennas, and 3D attitude can be determined using the geometry information from the GNSS antennas [12]. Here, we employed six single-frequency GNSS receivers to estimate the UAV's position and attitude. By using the "redundancy" of multiple GNSS receivers, we developed a novel technique that can enhance the positioning and attitude estimation accuracy in multipath environments such as those located in the vicinity of buildings. The proposed method does not use IMU to estimate the UAV's position and attitude, which, in turn, can reduce the cost of implementing the 3D mapping system as well as system complexity.

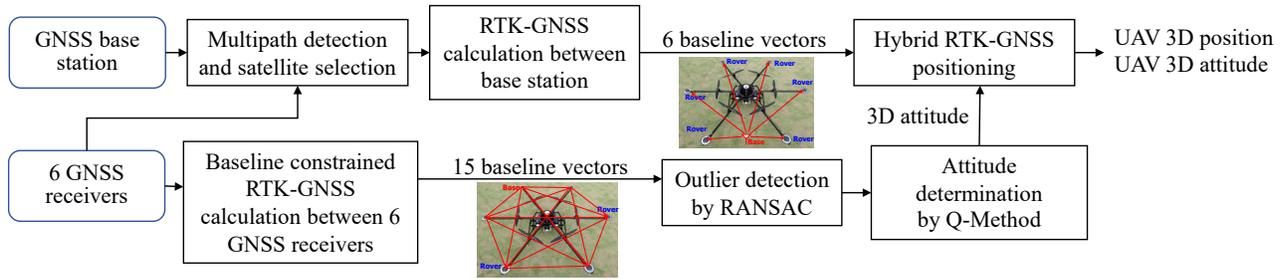

Figure 3. Overview of the proposed UAV position and attitude determination method by using multiple GNSS receivers/antennas.

Figure 3 depicts an overview of the proposed attitude and position estimation method. First, the UAV's 3D attitude is estimated using the six GNSS receivers/antennas; afterward, fifteen baseline vectors are estimated using the carrier-phase-difference GNSS technique. To improve attitude-determination accuracy in multipath environments, we used a baseline constraint to enhance the ambiguity-resolution performance of the single-frequency RTK-GNSS, and we used the random sampling consensus (RANSAC) method to detect and reject wrong fix solutions of the single-frequency RTK-GNSS. We propose the Q-Method, using which the optimal attitude can be directly determined from the fifteen baseline vectors. Subsequently, the UAV's 3D position is estimated using the known value of the UAV's 3D attitude obtained from hybrid RTK-GNSS positioning. The other method is performing the consistency examination of the GNSS's signal strength. Because various GNSS antennas involve different signal propagation paths, multipath fading occurs at varying times in the signal from each antenna/receiver. To examine the consistency between the signal-to-noise ratios (SNRs) of the GNSS receivers, the multipath signal can be determined. To reject the multipath signals, the fix rate of the RTK-GNSS can be enhanced. The remainder of this paper describes the details of the proposed methods to estimate a UAV's position and attitude.

## III. Attitude Estimation Using Multiple GNSS

To estimate the absolute attitude of the UAV, relative GNSS antenna positions (baseline vectors) across the six antennas/receivers determined using the moving-base RTK-GNSS technology, which is based on GNSS carrier-phase measurements, are used in this study. The difference of GNSS fix solutions of each antenna using a ground base station can also generate the baseline vectors between multiple GNSS antennas; however, to directly calculate the baseline vectors by using the moving-base RTK-GNSS technology, the accuracy of the baseline vectors can be improved to eliminate the common errors in the GNSS observations. Fifteen baseline vectors between the six GNSS antennas are depicted in Figure 4. In this process, an external ground GNSS base station is not required. One of the six GNSS receivers is treated as the base station. We use the Q-Method [13] to estimate the UAV's attitude from multiple baseline vectors. To determine the attitude of the GNSS, at least two baseline vectors must be estimated (the UAV's attitude can be estimated using these GNSS antennas). In this study, we have used fifteen baseline vectors to estimate the UAV's attitude to improve the efficiency of the single-frequency RTK-GNSS technology. The Q-Method is based on an optimization technique to obtain a satellite's attitude. The attitude representation is parameterized using the quaternions. Using the Q-Method, the optimal attitude can be directly determined from the fifteen baseline vectors, and we can improve the robustness of attitude determination.

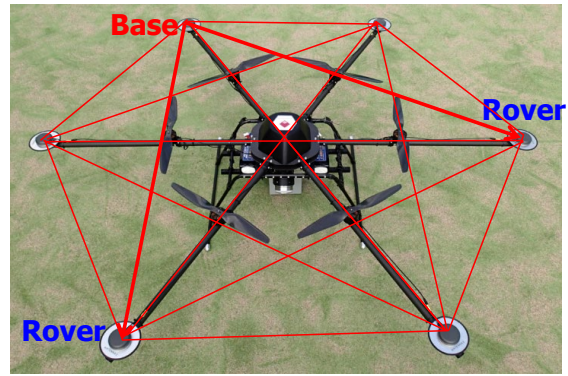

Figure 4. Baseline vectors computed from RTK-GNSS between multiple GNSS antennas for UAV's attitude estimation.

The Q-Method calculates the maximum-likelihood UAV attitude by calculating an attitude matrix that minimizes the errors of observation vectors to reference observation vectors. First, we define an East-North-Up (ENU) coordinate system, whose origin is a base station, and a body center body fixed (BCBF) coordinate system, whose origin is the center of the UAV. Furthermore, we define baseline vectors between the GNSS antennas in the ENU coordinate system as an observation vector $\mathbf{v}_i$ and in the BCBF coordinate system as a reference observation vector $\mathbf{w}_i$. From these vectors, we calculate the maximum-likelihood attitude matrix by calculating an orthogonal matrix $\mathbf{R}$ that minimizes a loss function $L(\mathbf{R})$, as shown in equation (1).

$$L(\mathbf{R}) = \frac{1}{2}\sum_{i=1}^{n} a_i |\mathbf{w}_i - \mathbf{R}\,\mathbf{v}_i|^2 \qquad (1)$$

$$\sum_{i=1}^{n} a_i = 1 \qquad (2)$$

where $a_i$ is for weighting and is weighted in proportion to the length of the baseline vector. In the Q-Method, consider the attitude matrix $\mathbf{R}$ that maximizes a gain function $g(\mathbf{R})$ as shown in equation (3), instead of minimizing the lost function

$L(\mathbf{R})$, as shown in equation (1). This results in the following eigenvalue problem.

$$g(\mathbf{R}) = 1 - L(\mathbf{R}) = \mathbf{q}^T \mathbf{K} \mathbf{q} = \lambda \quad (3)$$

$$\mathbf{K}\mathbf{q} = \lambda \mathbf{q} \quad (4)$$

Equation (4) represents an eigenvalue equation about a four-dimensional matrix $\mathbf{K}$, where $\mathbf{K}$ is an orthogonal matrix. Accordingly, all the eigenvalues are real numbers. In addition, equation (3) should be maximized. Therefore, the maximum eigenvalue among the four eigenvalues will be the likelihood eigenvalue. From the above, while using the Q-Method, we can calculate the most suitable quaternion $\mathbf{q}$ for the observation vectors by calculating an eigenvalue vector corresponding to the maximum eigenvalue in the eigenvalue equation (4). Using this method, the UAV's attitude can be estimated by employing multiple GNSS receivers.

In multipath environments, wrong ambiguity will sometimes be estimated. As a result, wrong baseline vectors will be estimated from single-frequency RTK-GNSS. In this situation, it is important to detect and reject the wrong baseline vectors from attitude computation. Here, we use the RANSAC-based algorithm to exclude the outlier of the baseline vectors. The RANSAC method is an iterative method for estimating a mathematical model from a data set that contains outliers. The UAV's attitude can be estimated using at least two baseline vectors; we randomly select a small subset of baseline vectors and estimate the UAV's attitude using the Q-Method. Moreover, we evaluate the error residuals for the rest of the measurements under the attitude estimated using randomly selected baseline vectors. Repeating this procedure, the outlier of the baseline vectors can be rejected. Using the proposed method, we can improve attitude-determination availability and accuracy in multipath environments.

## IV. Position Estimation using Multiple GNSS

The first approach combines the GNSS fix solutions of multiple GNSS antennas to enhance the fix rate of the GNSS carrier-phase ambiguities. This concept is highly straightforward while being highly effective in enhancing positioning availability and accuracy. In multipath environments, the carrier-phase multipath affects the ambiguity resolution of the GNSS. Because each GNSS antenna involves separate GNSS signal-propagation paths, each GNSS receiver/antenna exhibits divergent multipath errors. In general, only one GNSS receiver is used to estimate the UAV's position. If a multipath error occurs and if the carrier-phase ambiguities cannot be solved in the RTK-GNSS, then the determined UAV's position will be inaccurate. In contrast, the use of multiple GNSS receivers can enhance the probability of obtaining an RTK-GNSS fix solution because the UAV's position can be determined from the fix solution of at least one GNSS receiver. To combine multiple fix solutions, obtaining the accurate 3D attitude position of the UAV is necessary in the coordinate transformation process. The 3D attitude position of the UAV is estimated using multiple GNSS receivers, as described in the previous section. Figure 5 illustrates the outline of the proposed method. We use a GNSS base station to compute at most six RTK-GNSS fix solutions.

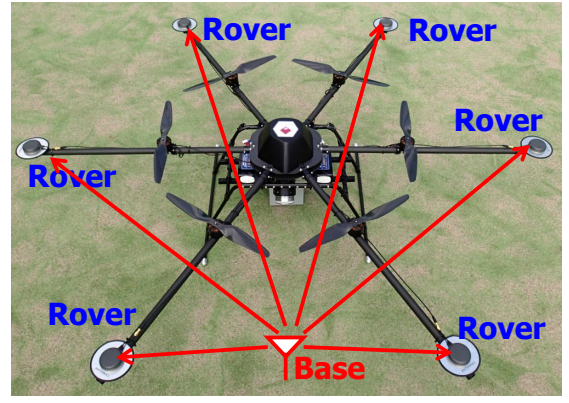

Figure 5. Hybrid positioning by using multiple GNSS receivers/antennas.

The proposed method is executed by following the below procedure: Assuming that $N$ RTK-GNSS fix solutions —solved by the ambiguity of the carrier phase—are obtained, the 3D position in the ENU coordinate system is

$$\mathbf{p}_i = \begin{bmatrix} x_i^{fix} & y_i^{fix} & z_i^{fix} \end{bmatrix}^T \quad (5)$$

where $i$ is the antenna number; in this study, $N$ is 6.

The obtained antenna positions are converted to the origin of the UAV coordinate system, which is fixed to the UAV. A vector of each GNSS antenna in the UAV coordinate system is

$$\mathbf{b}_i = \begin{bmatrix} x_i^b & y_i^b & z_i^b \end{bmatrix}^T \quad (6)$$

Because the geometric arrangement of the GNSS antennas is fixed, the antenna position $\mathbf{b}$ in the UAV coordinate system can be measured in advance. Assuming that the rotation matrix that represents the UAV's attitude from the UAV coordinate system to the ENU coordinate system is $\mathbf{R}$, the UAV's position in the ENU coordinate system according to the proposed hybrid positioning is as follows:

$$\mathbf{p} = \frac{\sum_{i=1}^{N}(\mathbf{p}_i - \mathbf{R}\mathbf{b}_i)}{N} \quad (7)$$

Using this method, if one fix solution is obtained from any one of the six GNSS receivers, it is feasible to calculate the position of the UAV with high accuracy. Compared with the case of using only one GNSS receiver, we can enhance the availability of carrier-phase ambiguity resolutions by using multiple single-frequency GNSS receivers (in this paper, we have used sic GNSS receivers/antennas). In addition, as the UAV's position is calculated from the average fix solutions when multiple fix solutions are obtained, it is likely that the positioning accuracy will be enhanced as compared with that in the case of using only one GNSS receiver. Thus, the total availability and accuracy of the positioning solutions can be enhanced in multipath environments also.

The second approach involves the detection and elimination of multipath signals by using the observations of multiple GNSS receivers. To further enhance the fix rate of the RTK-GNSS in an environment where multipath occurs, it is necessary to detect the satellites that face multipath errors. If a GNSS antenna receives a multipath signal from a satellite, the carrier-phase ambiguity is not resolved on a few occasions because of the multipath error. It is, therefore, essential to establish a positioning technique that identifies and selects the satellites that provide the lowest multipath error. We focus on the SNR of the multipath signal. In a multipath environment, the GNSS antenna simultaneously receives direct signals, along with multiple reflected and diffracted signals. As a result, the SNR of the combined received signals vibrates in the time direction. This phenomenon is called multipath fading. Because signals from various GNSS antennas follow divergent signal propagation paths, multipath fading occurs at separate times for each antenna signal. To examine the consistency between the SNRs of multiple GNSS receivers, the multipath signal is first determined. If all the antennas are in an open-sky environment, an identical SNR is likely to be observed for each GNSS antenna. If an antenna receives multipath signals, divergent SNRs are observed in each GNSS antenna. This is used to detect multipath signals.

Our algorithm is executed as follows: For each GNSS satellite, the SNR is obtained from the multiple GNSS receivers simultaneously. We calculate the standard deviation (SD) of the SNR as an index of SNR variation among multiple antennas.

$$\sigma_{SNR} = \sqrt{\frac{1}{N}\sum_{i=1}^{N}(SNR_i - \mu_{SNR})^2} \qquad (8)$$

where $SNR_i$ denotes the SNR of the $i^{th}$ GNSS antenna, $\mu_{SNR}$ denotes the average value of SNR, and $N = 6$ in this study. A large SNR SD indicates a high possibility of the occurrence of multipath signals. Therefore, GNSS satellites' signals with a large value of SNR SD are detected as multipath signal by a straightforward threshold test.

## V. STATIC EXPERIMENT

### A. Experimental Environment

To evaluate the proposed method, we conducted a static test in a narrow-sky environment that contained obstacles such as buildings, as illustrated in Figure 6. A fish-eye image captured at the GNSS antenna's location is illustrated in Figure 6. We used GPS, BeiDou, and QZSS to estimate the position of the UAV. The GNSS data were collected at 10 Hz. The reference station for RTK-GNSS was installed in the open-sky near the experimental location. In addition, the collected GNSS data were analyzed during post-processing.

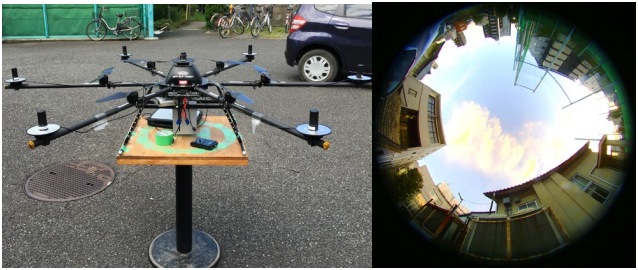

Figure 6. Test environment for position estimation in the static test and fish-eye image captured during the static test.

### B. Attitude-Estimation Results

We determined the UAV's attitude by using the proposed method. We compared the attitude's position accuracy and availability using the six GNSS antennas/receivers. Figure 7 depicts the UAV's attitude-estimation results. Table 1 presents the SD and availability of the attitude determination by the proposed method. Using the proposed method, we could almost perfectly solve the GNSS carrier-phase ambiguity by using low-cost single-frequency GNSS receivers in multipath environments. The availability of the UAV's attitude estimation was enhanced from 71.2% to 96.2% using the proposed method. The estimated roll, pitch, and yaw angle errors were approximately 0.1° upon using the six GNSS receivers. It can be concluded that the proposed method can improve the accuracy and availability of UAV's attitude estimation in multipath environments.

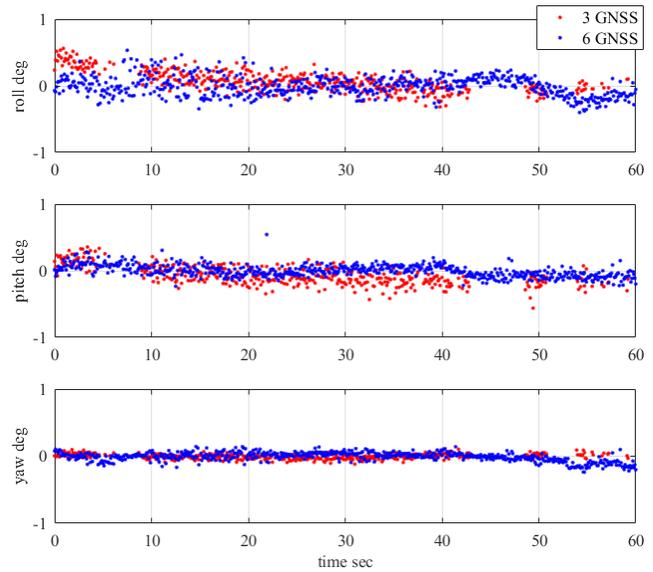

Figure 7. Attitude-estimation result in narrow-sky environment.

TABLE I. ACCURACY AND AVAILABILITY OF ATTITUDE DETERMINATION

|  | Roll SD deg | Pitch SD deg | Yaw SD deg | Availability % |
|---|---|---|---|---|
| 3 GNSSs | 0.182 | 0.162 | 0.041 | 71.2 |
| 6 GNSSs | 0.202 | 0.174 | 0.066 | 96.2 |

### C. Positioning Results

We evaluated the UAV's position estimation by using the proposed hybrid GNSS fix solutions. Figure 8 illustrates the number of receiving satellites for each antenna and the time when the RTK-GNSS fix solution was obtained. The points indicated by green in Figure 8 represent the time when the ambiguity of the carrier phase is solved by the RTK-GNSS and the fix solution is calculated. As illustrated in Figure 8, notwithstanding the same set of antennas, receiver, and RTK-GNSS algorithm being used, the fix rate and fix solutions obtained from each GNSS receiver/antenna can vary. The timing at which the fix solution can be obtained varies substantially depending upon each GNSS receiver. Table 2 presents the fix rate calculated using the conventional method and that calculated using the proposed hybrid positioning method. The fix rate of each GNSS receiver exhibited significant variation (17–68%). However, using the proposed hybrid positioning method, a high fix-rate of 94.3% was obtained by combining all the fix solutions.

To further enhance the fix rate, multipath detection and rejection are evaluated using the proposed method. The fix rate calculated using the proposed hybrid positioning method with multipath detection was enhanced from 94.3% to 97.3%. Moreover, the proposed method offers increased positioning accuracy in urban environments. The SDs of the fix solution obtained using the proposed hybrid positioning method with multipath detection were observed to be 5.3, 4.8, and 11.1 mm in the $x$, $y$, and $z$ directions, respectively. It was also verified that the position of a UAV can be estimated with high accuracy in an environment that contains buildings. Thus, by using multiple GNSS antennas/receivers, the proposed method can substantially enhance the fix rate of RTK-GNSS and estimate the UAV's position with high accuracy, also in an environment where multipath propagation occurs that decreases the UAV's positioning accuracy.

TABLE II. RTK-GNSS FIX RATE AND ACCURACY OF HYBRID POSITIONING

| Antenna Number | Fix rate | Fix rate (Hybrid positioning) | Fix rate (with multipath detection) | SD of the estimated position |
|---|---|---|---|---|
| 1 | 62.5 % | 94.3 % | 97.3 % | East; 5.3 mm North: 4.8 mm Up: 11.1 mm |
| 2 | 17.2 % | | | |
| 3 | 55.5 % | | | |
| 4 | 30.2 % | | | |
| 5 | 68.2 % | | | |
| 6 | 49.3 % | | | |

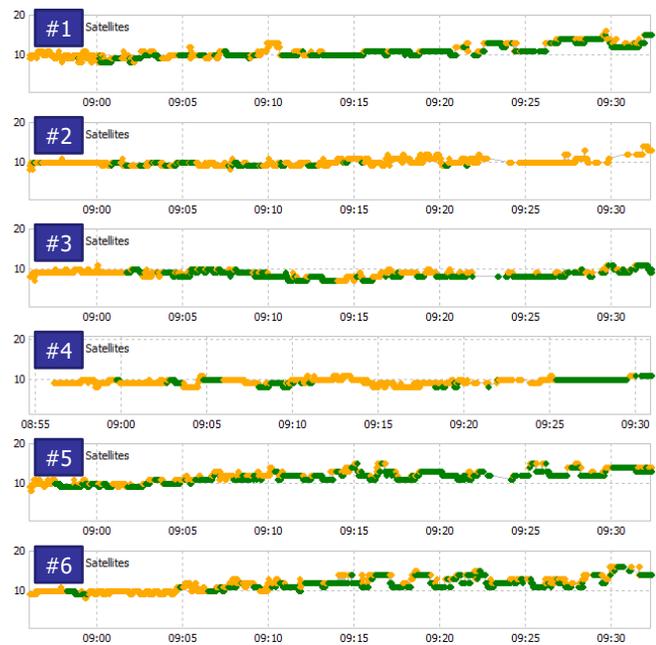

Figure 8. Fix rate of each GNSS receiver/antenna.

## VI. FLIGHT EXPERIMENT

### A. Experimental Environment

In the case of UAV's flight experiment, it is challenging to determine a reference position to evaluate the proposed method because a high-grade system that estimates the UAV's position and attitude cannot be installed in a small-sized UAV, owing to payload-related restrictions already mentioned. Therefore, we evaluated the proposed method to verify the accuracy of the 3D mapping. In this test, we installed several reflectors that served as GCPs in the test environment. Figure 9 illustrates the test environment and the locations of the installed reflectors. The locations of the reflectors were measured in advance by conducting the GNSS survey. The UAV's flight altitude in this experiment was approximately 30 m and flight speed was 3 m/s.

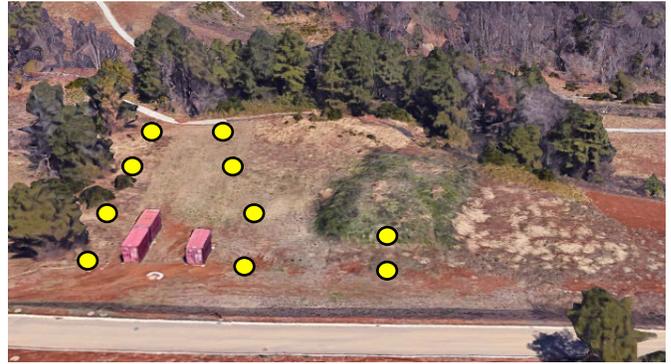

Figure 9. Test environment and reflectors' locations in flight test.

### B. 3D-Mapping Result

The estimated position of the UAV calculated using the six GNSS antennas/receivers is illustrated in Figure 10 (a). Each UAV position calculated by the GNSS antennas is transformed to UAV coordinates, and the UAV position is calculated using the proposed method (see Figure 10 (b)). The fix rate calculated using the proposed hybrid positioning method with multipath detection was observed to be 100%. Figure 11 illustrates the result of the 3D mapping. We evaluated the 3D measurement error due to the installed reflectors. The average root mean square (RMS) error in the horizontal position measurement was 4.7 cm, and that for the vertical position measurement 4.1 cm. Thus, we achieved 5 cm 3D-mapping accuracy using multiple GNSS receivers and laser scanner, and, that too, without using any other sensor.

## VII. CONCLUSION

Direct 3D mapping using a small-sized UAV equipped with a laser scanner is required in numerous remote-sensing applications. In direct 3D mapping approach, the precise estimation of the position and attitude of a UAV are necessary to construct accurate 3D maps. Because a small-sized UAV suffers from payload-related limitation, it becomes challenging to implement dual-frequency GNSS receivers and high-grade IMUs for 3D mapping. In this study, we proposed a precise UAV position and attitude estimation technique that employs multiple low-cost and light-weight GNSS antennas/receivers for small-sized UAVs, within the payload capacity of the UAV. Using the "redundancy" of multiple GNSS receivers, we can enhance the UAV position and attitude estimation accuracy in multipath environments such as those located in the vicinity of buildings. We developed a prototype UAV that was equipped with six GNSS antennas/receivers. From the experimental test results, we conclude that the proposed technique can enhance the accuracy of UAV's position estimation. Moreover, from the 3D mapping test, we conclude that the proposed system can measure 3D coordinates with 5 cm accuracy.

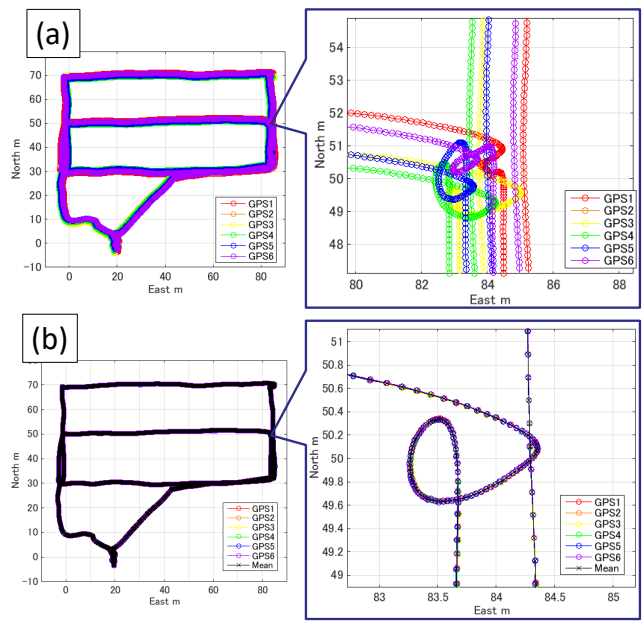

Figure 10. UAV trajectory estimated from multiple GNSS receivers.

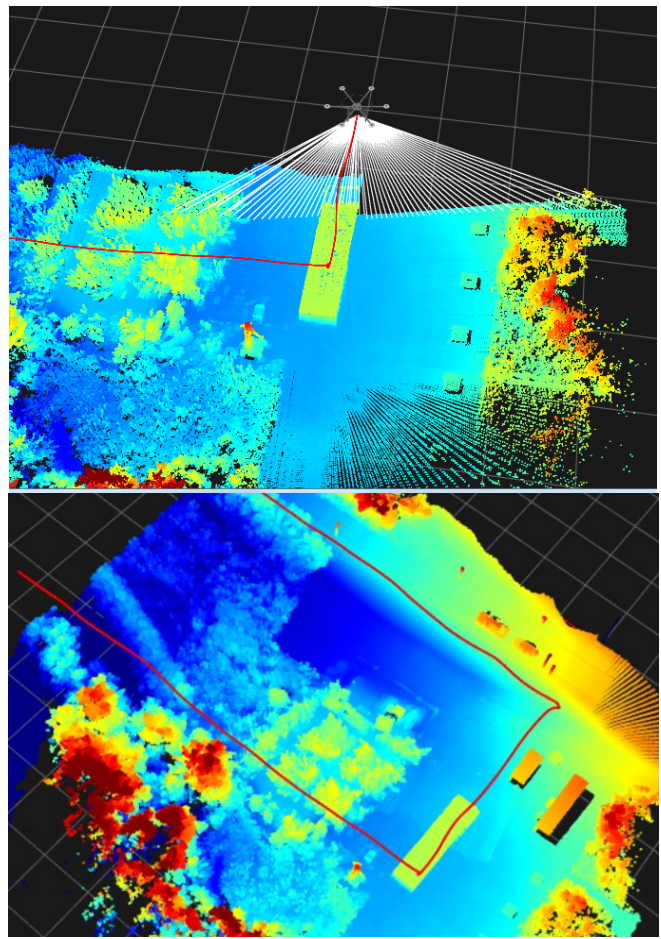

Figure 11. 3D-mapping result. The red line indicates the estimated trajectory of the UAV.